\documentclass[10pt, a4paper]{article}
\usepackage{lrec}
\usepackage{multibib}
\newcites{languageresource}{Language Resources}
\usepackage{graphicx}
\usepackage{tabularx}
\usepackage{soul}

\usepackage{epstopdf}
\usepackage[latin1]{inputenc}

\usepackage{xstring}

\usepackage{url}

\usepackage{color}
\usepackage{booktabs}
\usepackage{ctable}
\usepackage{xspace}
\usepackage{makecell}
\usepackage{graphicx}
\usepackage{amssymb}

\newcommand{\co}[1]{{\tt #1}\xspace}
\newcommand{\noconn}{\co{[No connective]}}
\newcommand{\da}{{\sc DA}\xspace}
\newcommand{\wordpairs}{{\sc WordPairs}\xspace}
\newcommand{\random}{{\sc Random}\xspace}
\newcommand{\Fone}{F\textsubscript{1}\xspace}

\title{Automatic Prediction of Discourse Connectives}

\name{Eric Malmi\textsuperscript{\normalfont{1},*\thanks{\textsuperscript{*}Work performed 
during an internship at Google.}},
Daniele Pighin\textsuperscript{\normalfont{2}},
Sebastian Krause\textsuperscript{\normalfont{2},*},
Mikhail Kozhevnikov\textsuperscript{\normalfont{2}}}

\address{%
\begin{tabular}{cc}
\textsuperscript{1}Aalto University & \textsuperscript{2}Google \\
Espoo, Finland & Z{\"u}rich, Switzerland \\
{\tt eric.malmi@aalto.fi} & {\tt \{biondo,bastik,qnan\}@google.com}
\end{tabular}}

\abstract{
	Accurate prediction of suitable discourse connectives (\emph{however}, 
	\emph{furthermore}, etc.) is a  key component of any system aimed at 
	building 
	coherent and fluent discourses from shorter sentences and passages. As an 
	example, a dialog system might assemble a long and informative answer by 
	sampling passages extracted from different documents retrieved from the 
	Web. 
	We formulate the task of discourse connective prediction and release a 
	dataset of 2.9M sentence pairs separated by discourse connectives for this 
	task.
	Then, we evaluate the hardness of the task for human raters, apply a 
	recently proposed decomposable attention (DA) model to this task and 
	observe that the automatic predictor has a higher \Fone than human raters 
	(32 vs. 30).  Nevertheless, under specific conditions the raters still 
	outperform the DA model, suggesting that there is headroom for future 
	improvements. \\ \newline 
}

\begin{document}

\maketitleabstract

\section{Introduction}

Discourse connectives, also referred to as discourse markers, discourse cues, or discourse adverbials, are used to bind together and to explicate the relation between pieces of text.
It is a common language class exercise to be asked to fill in suitable connectives to a text in order to improve the text flow. Similarly, it is important for computational summarization and text adaptation systems 
to be able to fill in suitable discourse connectives to produce natural-sounding utterances.

In this work, we study the problem of automatic discourse connective prediction. We limit ourselves to connectives which appear at the beginning of a sentence, linking the sentence to the preceding one. Even in this limited setting, an automatic discourse connective predictor has many concrete use cases. For example, in a question-answering setting it could help to generate answers by collating sentences from multiple sources. In extractive text summarization, it could be used to determine what is the best way to join two sentences that used to be separated by one or more sentences. As part of a text-authoring application, it could suggest suitable connectives at the beginning of a sentence.

In the literature, discourse connective prediction has been recently studied merely as an intermediate step for the well-studied problem of implicit discourse relation prediction~\cite{xu2012,zhou2010}. However, considering the aforementioned applications, we argue that connective prediction makes an interesting and relevant problem in its own right.

The contributions of this work are twofold:
\begin{enumerate}
\item We present an extensive experimental study on the problem of discourse connective prediction and show that a recently proposed decomposable attention model 
  \cite{parikh2016} yields a good performance on this task. The model clearly outperforms a popular word-pair model and 
  obtains a better performance than human raters on the same task and data.
\item We describe the dataset that we collected, consisting of 2.9 million adjacent sentence pairs (with and without a connective) extracted from the English Wikipedia. For 10\,000 sentences, we also include connectives filled in by human raters. The dataset is publicly available at: \url{https://github.com/ekQ/discourse-connectives}
\end{enumerate}

\section{Related Work}
A few earlier works study discourse connective prediction alone, but recently it has been studied merely as an intermediate step for discourse relation prediction. Next we provide a brief overview of these two lines of work, starting from the latter.

\subsection{Predicting Connectives for Implicit Discourse Relation Prediction}

Implicit discourse relation prediction has attracted considerable attention in recent years \cite{braud2016,liu2016,qin2016,qin2017,rutherford2015,wu2017,zhang2016}.
Earlier 
\newcite{pitler2008} showed that if a discourse connective is known, the explicit discourse relation\footnote{Later, it has been shown that a single discourse connective can actually convey multiple discourse relations~\cite{rohde2015,rohde2016}.} can be inferred with a 93.09\% accuracy, which has inspired several efforts at predicting connectives to improve implicit discourse relation prediction. \newcite{zhou2010} predicted connectives using an $N$-gram language model, 
whereas \newcite{xu2012} employed word pairs and a selection of linguistically informed features. \newcite{liu2016b}, on the other hand, showed that predicting both connectives and relations using a convolutional neural network in a multi-task setting improves the relation prediction performance.

\subsection{Discourse Connective Prediction}

Some earlier works have focused on connective prediction alone and developed various hand-crafted features for distinguishing between connectives. For example, \newcite{elhadad1990} explored pragmatic features for distinguishing between the connectives \co{but} and \co{although}, and between \co{because} and \co{since}. Later \newcite{grote1998} developed a specialized lexicon for discourse connectives based on the relevant constraints and preferences associated with the connectives. While these works do not present an experimental evaluation of the proposed systems, we evaluate our connective prediction models extensively in order to understand their applicability to real-life scenarios. Furthermore, our aim is to learn the representations of the two arguments and their relationship automatically which allows us to distinguish between a large set of connectives without extensive manual efforts required to craft features that separate the connectives.

In addition to predicting the most suitable discourse connective, several methods have been developed for predicting the presence of a discourse connective~\cite{yung2017,di1997,patterson2013}. We also predict the presence of a connective by considering \noconn as one of the classes to be predicted.

\section{Data Collection}
\label{sec:data}

We compile a list of 79 discourse connectives based on the Penn Discourse Treebank (PDTB)~\cite{prasad2008}. Since our focus is on sentence concatenation, we ignore the (forward) connectives, which typically point to the following sentence rather than the previous one, such as ``\co{After} the election, [\ldots]''. However, for several ambiguous connectives, the forward use can be ruled out by requiring a comma after the connective (e.g. \co{Instead,}); we include such connectives in our data. Discontinuous connectives, such as ``\co{If} [\ldots] \co{then} [\ldots]'', are not included.

Data samples for discourse connective prediction can be collected from any large unannotated text corpus. In this instance, we use the English Wikipedia\footnote{A snapshot from September 5, 2016.} and collect every pair of consecutive sentences within the same paragraph where the latter sentence begins with one of the 79 discourse connectives. As a result, we obtain a dataset of 1.95 million sentence pairs separated by a connective. Additionally, we collect 0.91 million examples of consecutive sentences not separated by a discourse connective, labeled as \noconn, for a total of 2.86 million sentence pairs.\footnote{Note that our models are tested only on consecutive sentences, for which the ground truth connectives are known, but they can be applied to connect also disjoint sentences.}

The frequency distribution of the connectives is very skewed; \co{however} occurs 720\,334 times, whereas \co{else,} only 43 times in the beginning of a sentence. 
In order to make the connective prediction task more feasible for the models and for human raters,
we select a subset of sufficiently frequent and distinct connectives 
(e.g. \co{for example} is included but \co{for instance} is not since it conveys the same meaning and is less frequent). The details of the selection process are omitted in the interest of space, but the resulting 19 connectives are listed in Table~\ref{tab:19}. 

\begin{table}[tb]
\centering
\resizebox{\columnwidth}{!}{%
\begin{tabular}{rrrr}
\toprule
 Connective & Occurrences & Connective & Occurrences \\
\midrule
\co{however} & 720\,334 & \co{on the other hand} & 20\,301 \\
\co{for example} & 111\,711 & \co{in particular,} & 16\,011 \\
\co{and} & 73\,644 & \co{indeed,} & 15\,286 \\
\co{meanwhile,} & 57\,971 & \co{overall,} & 9\,513 \\
\co{therefore} & 44\,064 & \co{in other words} & 8\,888 \\
\co{finally,} & 33\,076 & \co{rather,} & 5\,596 \\
\co{nevertheless} & 32\,952 & \co{by contrast,} & 4\,605 \\
\co{instead,} & 30\,973 & \co{by then} & 4\,279 \\
\co{moreover} & 25\,583 & \co{otherwise,} & 3\,563 \\
\co{then,} & 21\,731 &  &  \\
\bottomrule
\end{tabular}
}
\caption{The list of 19 connectives studied in the experiments in addition to the \noconn class. Only the connectives which are sufficiently frequent and distinct in their meaning have been selected.}\label{tab:19}
\end{table}

Finally, we split the data into train, development, and test sets. We balance the connective classes, since in an unbalanced dataset most examples would be labeled as \noconn and many connectives would be extremely under-represented, limiting the applicability of the resulting classifier. For the development and test sets, we pick 500 samples per connective (including \noconn) by under-sampling without replacement. This results in two balanced datasets of 10\,000 samples. For the training set, we pick 20\,000 samples per connective by under-sampling the majority classes and oversampling the minority classes, creating a balanced dataset of 400\,000 samples. Connective samples from a single Wikipedia article are not included in more than one of the three datasets to avoid over-fitting through potential repetition within a single article.

In comparison with the PDTB dataset, which contains information about both discourse connectives and discourse relations, the main advantage of the collected dataset is its size. PDTB contains only 40\,600 examples (1.4\% of the size of the collected dataset), which causes sparsity issues~\cite{li2014}. This can slow down the development of new models, particularly complex neural models that often require large training datasets to generalize well.

\section{Connective Prediction Models}
\label{sec:method}

The decomposable attention (\da) model was recently introduced by 
\newcite{parikh2016} for the \textit{natural language inference} (NLI) problem 
which aims to classify entailment and contradiction relations between a premise 
and a hypothesis. Discourse connective prediction is related to the NLI problem 
since entailment and contradiction can be explicitly indicated by certain 
connectives (for instance, \co{therefore} and \co{by contrast}, respectively). 
However, the larger number of classes makes connective prediction 
more challenging. 
\da was shown to yield a state-of-the-art performance on the 
NLI task while requiring almost an order of magnitude fewer parameters than 
previous approaches. For all these reasons, it seems natural to apply the \da 
model to the connective prediction problem.

\newcite{marcu2002} proposed to use word-pair features to predict discourse relations based on discourse connectives mapped to these relations. Similarly, many later implicit discourse relation prediction models are based on word-pair features \cite{marcu2002,pitler2009,xu2012,zhou2010} or aggregated word-pair features \cite{biran2013,rutherford2014}. Therefore, we use a model called \wordpairs to have a baseline for the \da model.

\subsection{The Decomposable Attention Model}

The \da model consists of three steps, \textit{attend}, \textit{compare}, and \textit{aggregate}, which are executed by three different feed-forward neural networks $F$, $G$, and $H$, respectively. 
As input, the model takes two sentences $\mathbf{a}$ and $\mathbf{b}$ represented by sequences of word embeddings. The sequences are padded by ``NULL'' tokens to fix their lengths to 50 tokens.

In the {\bf attend} step, 
the model computes non-negative attention scores for each pair of tokens across the two input sentences. This computation ignores the order of the tokens and it produces soft-alignments from $\mathbf{a}$ to $\mathbf{b}$ and \emph{vice versa}.

In the {\bf compare} step, 
the model computes comparison vectors between each input token and its aligned sub-phrase. The aligned sub-phrase is a linear combination of the embedding vectors of the other sentence weighted by the attention scores.

Finally, in the {\bf aggregate} step, the comparison vectors are summed over the tokens of a sentence and then the aggregate vectors of the two sentences are concatenated. The resulting vector is fed into 
the third feed-forward network which outputs $\hat{\mathbf{y}}$ containing scores for each class. The predicted class is given by $\hat{y} = \arg\max_i \hat{y}_i$.

The weights of the three networks are randomly initialized, after which the model is trained in an end-to-end manner. Our implementation of the \da model has the following differences compared to the original model described by \newcite{parikh2016}: ($i$)~we do not use the self-attention mechanism which was reported to provide only a small improvement over the vanilla version of \da; ($ii$)~we do not project down the embedding vectors but use 100-dimensional word2vec embeddings \cite{mikolov2013} which are updated during the training; 
($iii$)~we use layer normalization \cite{ba2016} which makes the model converge faster.

\subsection{The Word-Pair Model}

The \wordpairs model considers as features all word pairs which appear across the two arguments (e.g. word $A$ appears in Arg 1 and word $B$ in Arg 2) in at least five samples in the training dataset. Such features are employed by many implicit discourse relation prediction models \cite{marcu2002,pitler2009,zhou2010,xu2012}. Additionally, we incorporate single word features (e.g. word \emph{A} appears in Arg 2) since these slightly improved the results. With these binary features, we train logistic regressors using the one-vs-rest scheme to predict one of the 20 different connectives.\footnote{We trained two versions of the \wordpairs model: using stochastic gradient descent with mini-batches and using LIBLINEAR with 100k samples (i.e. 25\% of the training data) which we could fit into the memory of a 256 GB machine. The reported results are based on the latter approach, which performed better.}

\section{Experiments}

Next we present experimental results on discourse connective prediction using human raters, the \da model and the \wordpairs model.
For this task,
we remove the connective (if any) from the second sentence in each test pair, and measure the ability of the model (or the raters) to identify the removed connective. 

\subsection{Accuracy of Human Raters}\label{sec:human_accuracy}

To better understand what is reasonable to expect from an automatic predictor, we use a crowd-sourcing platform to ask human raters to reconstruct the removed connectives for each of the 10\,000 test sentence pairs.
Each sentence pair is annotated by three (not necessarily the same three) native English speakers. The raters are shown the two 
sentences, the latter of which starts with a \textit{[Connective goes here]} 
placeholder, and asked to select the most suitable connective from the 20 
options, including \noconn. This layer of human annotations is also released as 
part of the connective dataset. The raters are instructed to pick the most 
natural connective in case there are multiple suitable options. Furthermore, 
they are asked to pick \noconn only if adding a connective would make the 
concatenation sound ungrammatical or artificial, or
if the two sentences seem to be completely disconnected. The sentences are not pre-processed apart from upper-casing the first character of the second sentence to avoid giving away the presence of a connective in the original sentence. The order of the connectives is randomized, except for \noconn which is always shown last. 

\begin{figure*}[tb]
  \centering
  \includegraphics[width=\textwidth]{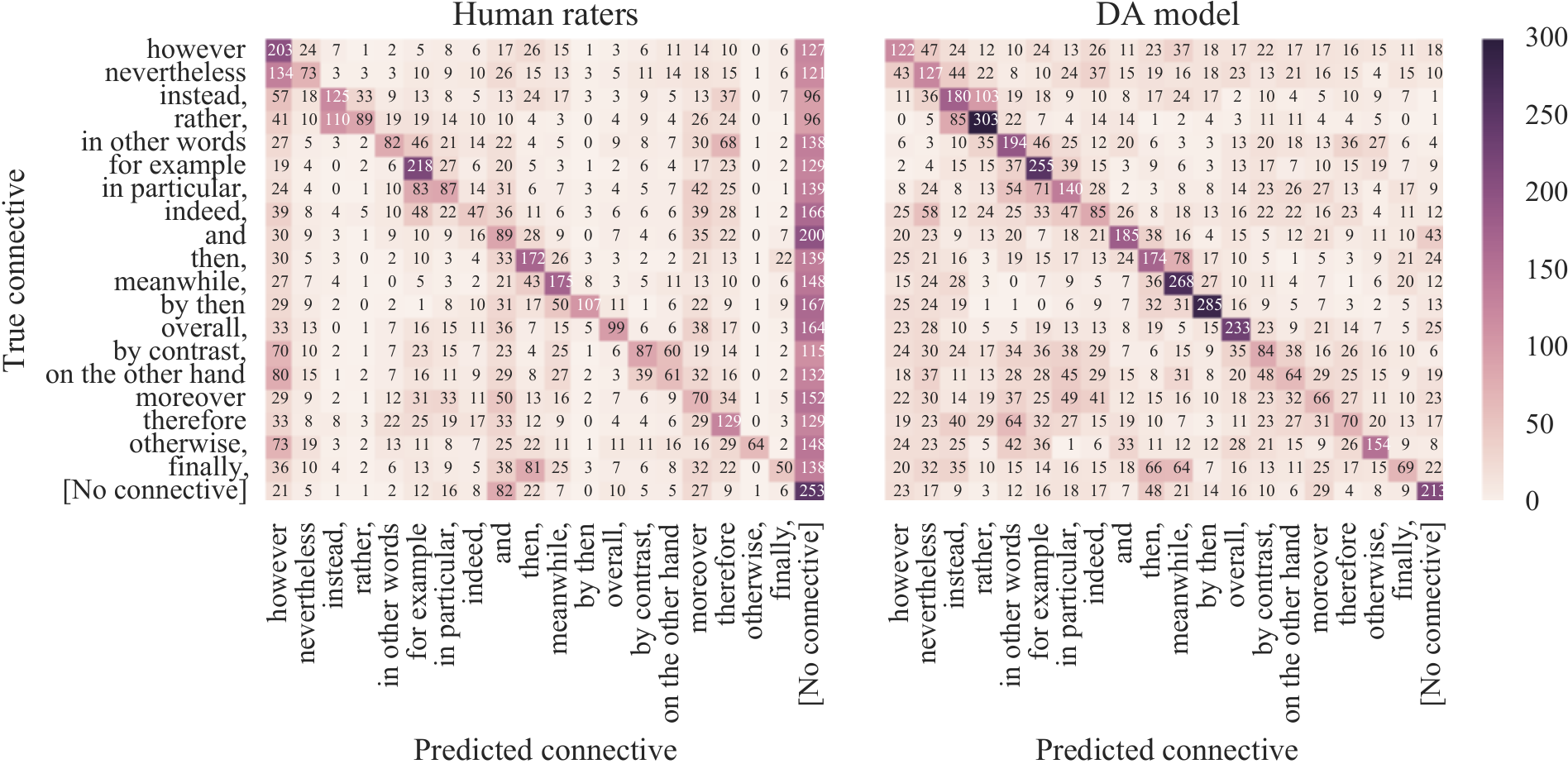}
  \caption{Confusion matrices for the human raters (left) and DA model (right) predicting the discourse connectives used by Wikipedia authors. The counts of the human raters are divided by three (i.e., the number of raters) for easier comparison.\label{fig:cm}}
\end{figure*}

On the whole test-set, human annotators achieve a macro-averaged \Fone score of 23.72. 
The confusion matrix generated by the raters' decisions is presented on the left side of Figure~\ref{fig:cm}. It shows that the raters are strongly biased towards \noconn despite the indication to refrain from using it. A similar bias was observed by \newcite{rohde2016} for the task of filling in a suitable conjunction before a discourse connective.
There are at least two possible explanations for this bias: ($i$) in the sake of clarity and in line with common scientific writing guidelines, Wikipedia editors tend to use connectives quite generously, and ($ii$) the artificial balancing of the datasets makes \noconn under-represented in the test data compared to the actual distribution of discourse connectives vs. \noconn.
The confusion matrix also shows that there are clusters of connectives that 
raters tend to confuse, even though they do not necessarily encode exactly the 
same relation. Examples are \co{rather,} and \co{instead,}, \co{for example} 
and \co{in particular,}, \co{on the other hand} and \co{by contrast,}.
For 57.1\% of the test questions, there is a consensus among at least two raters and for 11.4\%, all three raters agree on the most suitable connective.

\begin{table}[tb]
\centering
\begin{tabular}{ccc}
\toprule
 Model & Macro \Fone & Accuracy  \\
\midrule
\random & 5.00 & 5.00 \\
\wordpairs & 14.81 & 15.60 \\
\da & {\bf 31.80} & {\bf 32.71} \\
Human Raters & 23.72 & 23.12 \\
\bottomrule
\end{tabular}
\caption{Discourse connective prediction performance of a \random baseline, the \wordpairs model, the decomposable attention model (\da) and human raters.}\label{tab:results}
\end{table}

\subsection{Accuracy of the Models}

In this section, the \da model and the \wordpairs model are employed to perform the same task as the human raters, i.e., learning to reconstruct the connective possibly removed from the beginning of the second sentence in each test pair.
A balanced dataset is used for both training and testing the models as described in Section~\ref{sec:data} 
The \da model is evaluated using the following hyper-parameters optimized on the development set:
network size (one hidden layer with 200 neurons), batch size (64), dropout ratios for the $F$, $G$, and $H$ networks (0.68, 0.14, and 0.44, respectively), and learning rate (0.0018). The model is implemented in TensorFlow~\cite{tensorflow} and the training is run for 300\,000 batch steps.
The results, reported in Table~\ref{tab:results}, show that \da clearly outperforms the \wordpairs baseline with an \Fone score of 31.80 vs. 14.81.

\subsection{Comparison Between the Raters and the Decomposable Attention Model}

Table~\ref{tab:raters} compares the accuracy of \da predictions to the rater 
decisions. The macro-averaged \Fone score of human raters is 23.72 which is, quite 
surprisingly, lower than the \Fone score of the \da model, 31.80. The difference 
is smaller when considering majority votes on the subset of 5\,714 tasks for 
which 
there is a consensus among at least 2 out of 3 raters, which results in a 30.36 
\Fone score for the raters. On these less ambiguous cases, the model 
performance also increases to 32.68. 

\begin{table}[tb]
\centering
\begin{tabular}{cccc}
\toprule
 Setting & $n$ & Raters (\Fone) & Model (\Fone) \\
\midrule
A & 10\,000 & 23.72 & {\bf 31.80} \\
B & 5\,714 & 30.36 & {\bf 32.68} \\
C & 3\,204 & {\bf 41.97} & 36.65 \\
\bottomrule
\end{tabular}
\caption{Macro-averaged \Fone scores for human raters and the decomposable attention model. The three settings are: (A)~All test set items; (B)~Only the items for which there is a consensus among at least 2 out of 3 raters; (C)~Consensus items ignoring those where either ground truth, or the rater assigned, or the model assigned label is \noconn.}\label{tab:raters}
\end{table}

As we mentioned in Section~\ref{sec:human_accuracy}, human raters are clearly less eager to introduce a connective than Wikipedia editors. Therefore, we also evaluate the setting in which we exclude the questions for which either the ground-truth label, or the rater-assigned majority label, or the model-assigned label is \noconn. The results, listed in the last line of Table~\ref{tab:raters}, show that under these conditions human raters actually outperform the model.

\begin{figure*}[tb]
  \centering
  \includegraphics[width=\textwidth]{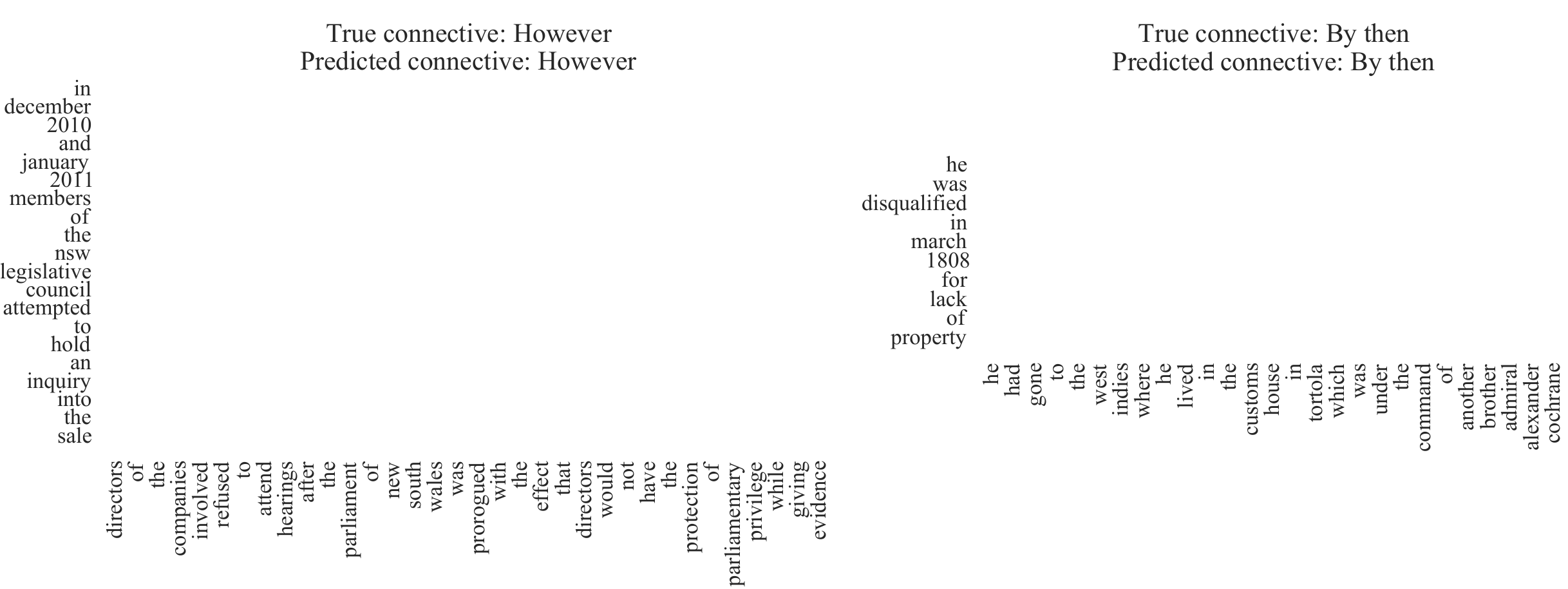}
  \caption{Two examples of the alignment matrix between the first ($y$-axis) and the second argument ($x$-axis) generated by the \da predictor. The darker the color, the higher the alignment score. \label{fig:attention1}}
\end{figure*}

The confusion matrix of \da is shown on the right side of Figure~\ref{fig:cm}. For each connective, the true connective is the most frequent prediction. Connective \co{on the other hand} has the lowest \Fone score (15.06), whereas \co{by then} has the highest (57.29). Some of the most frequent mistakes are between similar connectives, such as \co{however} vs.\ \co{nevertheless}, and \co{instead,} vs.\ \co{rather,}. These errors are by and large consistent with those of human raters (left side of the figure). This confirms that the model is accurately capturing the meaning of the relation, and when it does not select the gold connective it is making similar approximations to what people would do.
Furthermore, the Figure~\ref{fig:cm} shows that raters have a more pronounced tendency to select frequent connectives, such as \co{however} and \co{and}.
To further exemplify, in Table~\ref{tab:examples} we show a selection of wrong and correct decisions made by \da and human raters. 
A manual inspection of these and other examples shows that
in some cases a larger context than the previous sentence is required for inferring the connective. For instance, to correctly decide whether \co{finally,} is more suitable than \co{then,} one may have to inspect a larger context.

\begin{table*}[tb]
\centering
\resizebox{\textwidth}{!}{%
\begin{tabular}{cccc}
\toprule
Arguments & Gold & \da & Raters \\
\midrule
\makecell[l]{Arg 1: From 1913 to 1917 Lucey served as Illinois Attorney General. \\
Arg 2: Lucey was appointed to the Illinois Public Utilities Commission in 1917 and served until 1920.} & \co{\textbf{then}} & \co{\textbf{then}} & \noconn \\[1em]
\makecell[l]{Arg 1: In Ahmedgarh Municipal Council, Female Sex Ratio is of 901 against state average of 895. \\
Arg 2: Child Sex Ratio in Ahmedgarh is around 841 compared to Punjab state average of 846.} & \co{\textbf{moreover}} & \co{\textbf{moreover}} & \co{meanwhile} \\[1em]
\makecell[l]{Arg 1: If the question was answered correctly, the team would receive the next clue. \\
Arg 2: The chosen team member would have to try again.} & \co{\textbf{otherwise}} & \co{\textbf{otherwise}} & \co{then} \\[1em]
\makecell[l]{Arg 1: Lee then continued on to Boston, arriving 25 June. \\
Arg 2: The ranks of General Washington's Navy were being thinned by captures.} & \co{\textbf{meanwhile}} & \co{by then} & \co{\textbf{meanwhile}} \\[1em]
\makecell[l]{Arg 1: Cooper was promoted as an alternate leader to Ahern. \\
Arg 2: It was thought he could shore up the National Party's vote in its conservative rural heartland.} & \co{\textbf{in particular}} & \co{however} & \co{\textbf{in particular}} \\[1em]
\makecell[l]{Arg 1: Taylor urged Pius XII to explicitly condemn Nazi atrocities. \\
Arg 2: Pius XII spoke against the "evils of modern warfare", but did not go further.} & \co{\textbf{instead}} & \co{in particular} & \co{\textbf{instead}} \\
\bottomrule
\end{tabular}
}
\caption{Examples of mistakes and correct predictions made by the \da model and by the raters.}\label{tab:examples}
\end{table*}

\subsection{Model Interpretation}

An advantage of the \da model is that it is possible to examine which words the model attends to when inferring a connective. In some cases, the attended words are clearly meaningful semantically or linguistically, whereas in other cases the soft-alignment matrix that the model produces is harder to interpret. Examples of the former case are represented in Figure~\ref{fig:attention1}, which shows the alignment matrices from the tokens of the first sentence ($y$-axis) to the tokens of the second sentence ($x$-axis) so that the rows sum to 1.
In the left example, the model correctly predicts \co{however} as the connective after aligning the word \textit{attempt} with \textit{refuse} and \textit{not}. These word pairs indicate contrast which makes \co{however} a likely connective. In the right example, the model aligns the phrase \textit{was disqualified} with \textit{had gone} and correctly predicts \co{by then} as the connective. The corresponding tenses, i.e., past and past perfect, respectively, are likely clues of the presence of \co{by then}.

\section{Conclusions}

We studied the problem of discourse connective prediction, which has many useful applications in text summarization, adaptation and conversationalization. We collected a dataset of 2.9 million pairs of consecutive sentences and connectives, and made it publicly available to facilitate further research on this problem, as well as other related bi-sequence classification tasks. We showed that the recently proposed decomposable attention model performs surprisingly well on the connective prediction task, even better than human raters on the same representative test set consisting of 10\,000 samples. We also observed that, unlike the model, 
human raters have a preference for implicit connectives, as they do outperform the model if the comparison is restricted to the cases in which the majority of raters agrees on an explicit connective. The alignment matrices produced by the model suggest that the predictor is picking up relevant lexical, syntactic and semantic clues. The confusion matrix of the predictor shows very similar error patterns to the matrix generated from human raters, further confirming the meaningfulness of the decisions made by the model.

\section{Acknowledgements}

We would like to thank Cesar Ilharco for his help on running the experiments.

\section{Bibliographical References}
\label{main:ref}

\bibliographystyle{lrec}
\bibliography{references}

\end{document}